# Text Understanding in GPT-4 vs Humans

Thomas R. Shultz[1,2], Jamie M. Wise[1,2], Ardavan S. Nobandegani[1,3]

[1]Department of Psychology, McGill University

[2]School of Computer Science, McGill University

[3]Mila - Quebec AI Institute

We examine whether a leading AI system GPT-4 understands text as well as humans do, first using a well-established standardized test of discourse comprehension. On this test, GPT-4 performs slightly, but not statistically significantly, better than humans given the very high level of human performance. Both GPT-4 and humans make correct inferences about information that is not explicitly stated in the text, a critical test of understanding. Next, we use more difficult passages to determine whether that could allow larger differences between GPT-4 and humans. GPT-4 does considerably better on this more difficult text than do the high school and university students for whom these the text passages are designed, as admission tests of student reading comprehension. Deeper exploration of GPT-4's performance on material from one of these admission tests reveals generally accepted signatures of genuine understanding, namely generalization and inference.

1. Introduction

Recent advances in artificial intelligence (AI) have generated vigorous debates about whether these computational systems are approaching the level of Artificial General Intelligence (AGI), with humans often serving as the gold-standard of comparison. These computational systems are variously described as Large Language Models (LLMs) because they are large neural networks trained on enormous amounts of text, or Chat-Bots because it is possible to chat with them online, or generative AI because it can generate novel material. There is growing evidence that LLMs have mastered a variety of challenging human cognitive tasks involving language, mathematics, computer coding, law, medicine, vision, and more. However, there are also arguments that such systems that are trained to predict the next token word in vast quantities of sentences cannot reach the subtleties and power of human cognition and language, in some cases supported by adversarial inputs that manage to trip up some of the current AI systems (1,2).

An important review of the empirical literature on these debates has identified an interesting recent trend wherein newer, more powerful systems (e.g., GPT-4) have significantly improved on the shortcomings of earlier systems (e.g., GPT-3 and GPT-3.5) (3). The basic research strategy in these studies is to use an LLM to play the role of human participants in psychology experiments, by now covering an increasingly wide variety of challenging cognitive tasks. In such comparisons, sufficiently powerful LLMs often reached or even exceeded human levels of performance.

For example, GPT-3.5 and GPT-4 were tested on a range of analogical reasoning tasks, including one with the rule structure of Raven's Standard Progressive Matrices, a widely used test of human intelligence. In that IQ test, GPT-3.5 matched or exceeded average human performance and GPT-4 did even better (4). The authors noted that these models required no learning beyond their general training, an impressive demonstration of zero-shot learning.

Another example involved Theory-of-Mind tasks, which had often been considered exclusive to humans and thus impossible for LLMs to solve. Thus, it was not too surprising that early GPT models (3 and 3.5) solved only 20% of these tasks. However, the more powerful GPT-4 solved 75% of them, matching the performance of six-year-old children (5).

Other work reviewed in (3) showed that earlier LLMs performed at average human levels on a variety of challenging cognitive tasks, including metaphors, idioms, estimation, anaphoric reference, creativity, problem solving, logic, and answering common sense questions. GPT-3 also simulated average human results in gambles (including some human biases), and solved a multiarmed bandit task better than human participants (6). It would be interesting to see whether GPT-4 could exceed human performance on these tasks. It is noteworthy that causal reasoning (6), planning (7), and graduate-level mathematics (8) have been more challenging for LLMs, even for GPT-4.

In this work, we focus on the critically important issue of whether LLMs understand the text they have read, using standard psychological methods in which GPT-4 plays the role of a human agent. Section 2 deals with simple passages of text, while section 3 concerns considerably more difficult passages.

In psychology, text comprehension consists of building multi-level representations of the information in a passage of text (9). The comprehension improves when the reader has enough background knowledge to assimilate the text and as the reader constructs more representation levels and more inferences at each level. Successful comprehension can be measured by any of several abilities: correctly answering relevant questions, drawing relevant and correct inferences, asking good questions, generating good summaries of the text, and detecting anomalies and contradictions. Generalization is considered as a kind of inference that has fairly wide application. Knowledge-based inferences are constructed when background knowledge in long-term memory is activated and then encoded in the meaning representation of the text. We examine as many of these abilities as the data allow to assess text comprehension in humans and GPT-4.

2. Understanding Relatively Simple Passages

2.1 Methods

GPT-4 generates novel sentence content, has been pre-trained on vast amounts of unlabeled text, and uses a transformer architecture that leverages attention mechanisms to focus on relevant parts of sentences that may have difficult long-range dependencies. It has been recently trained by OpenAI researchers on over 45GB of language data processed by a large neural network with 1.76 trillion parameters (trainable connection weights). It is generally acknowledged to be the most powerful of the current LLMs.

The Discourse Comprehension Test (10) has several features that recommend its use for determining how well LLMs understand what they read: 1) It focuses entirely on how well verbal text is understood, 2) it is unknown to LLMs because it is protected for medical use, 3) it has been standardized on brain damaged patients known to have difficulty with text understanding as well as on neurotypical controls, and 4) its items are structured to experimentally examine the important variables of directness (stated vs. implied information) and salience (main idea vs. detail).

This test is comprised of 12 stories describing slightly humorous events that would be understandable to most North American adults. Each story contains between 191 and 210 words combined to create 13 or 14 sentences. The stories are at the fifth or sixth grade reading level, and are thus relatively easy for North American adults to understand (11). In the Discourse Comprehension Test, story comprehension is measured by eight yes/no questions characterized by salience (main idea vs. detail) and directness (stated vs. implied information).

There are two questions probing understanding of each of four distinct question types: stated main ideas, implied main ideas, stated details, and implied details, making a total of eight questions per story. Questions on the main idea concern central information that gets elaborated on by other information in the story. Questions on details concern peripheral information that is mentioned only once in the story. Stated questions use the same wording as in the story, while implied questions focus on information that is not directly stated but rather must be inferred from other information in the story. Answering implied questions correctly thus requires a participant to make bridging assumptions and draw inferences. An example story, along with its questions and scoring, is presented in Appendix A.

This test has been standardized on three groups of 20 brain-damaged patients (aphasia, right hemisphere brain damage, or traumatic brain injury) known to have difficulties comprehending discourse, as well as 40 adults without brain damage (12). Our focus is on comparing GPT-4 to these 40 neurotypical people. Participants in each of the four human groups were told five test stories after two non-scored practice stories. The three brain-damaged groups performed significantly worse than did the non-brain-damaged control participants.

It is very unlikely that GPT-4 has previously encountered any of the stories used in the Discourse Comprehension Test because this is a protected medical test in the field of Speech and Language Pathology, with stories and questions that are purposely kept out of the public eye and ear. Here we use 11 of these stories for testing GPT-4, leaving out the one story that uses true/false questions rather than yes/no questions. We ran each of the 11 stories through Copilot GPT-4 on 3 March 2024, preserving the answers given to each of the eight questions per story (10). Every answer was printed out well within the five seconds allowed for answers in the human experiment (12). An example of GPT-4's responses to the 8 questions for the story in Appendix A is presented in Appendix B. This story is chosen because it had already been posted as an example in an article describing a human study of discourse comprehension (12).

In our first experiment, we use two extra prompts for GPT-4. One prompt precedes the story: *Read this story in preparation for answering eight yes/no questions about the story*. The other prompt follows the story: *Answer each of these yes/no questions about the story*. Each story is itself a necessary prompt.

In a follow-up experiment run through Copilot GPT-4 on 2 April 2024, we instead use a prompt to summarize the story and mention main ideas not stated in the story: *Summarize this story, mentioning main ideas that are not stated and must be inferred*.

In our first experiment, we test GPT-4's ability to understand brief stories with yes/no questions structured to manipulate the salience and directness of parts of a story. Each of the 88 answers (8 answers per 11 stories) is categorized as *correct*, *wrong*, or *unsure*. An answer is correct if it matches the designated correct answer (*yes* or *no*) (10). Unlike the human participants, who apparently always conformed to answering only *yes* or *no* in their experiment (12), GPT-4 occasionally hedges by providing a neutral answer. Here is an exhaustive list of these neutral answers in our experiment: *The story does not specify …*, *not specified*, *not mentioned*, or *The* *story does not provide information on this*. For these hedged cases, we score the answer's correctness as .5 because it is approximately midway between correct (coded 1) and wrong (coded 0). None of these answers merits a score of 0, because each of the six incorrect answers are hedged; they are uncertain rather than being correct or wrong. For completeness, we also alternatively score hedged responses as 0, rather than .5.

2.2 Results

Because there are considerably more data points in the human sample (5 stories x 8 questions x 40 participants = 1600), we compare a single GPT-4 performance to human performance in terms of proportion of correct answers. Proportions correct in the human control sample are computed from Table 2 in the human study (12). Our Table 1 presents summary results for humans vs. GPT-4 with each of the two scoring methods for hedged responses. Although GPT-4 does very slightly better than humans for each of the two scoring methods, both differences are far below statistical significance. For the statistical tests in this section, we use the Two Sample Independent Proportions Test Calculator at Purdue University, a binomial test available online requiring input of sample size and successes for each of the two types of participants (humans and GPT-4).

Table 1: Comparison of two scoring methods for GPT-4 to human proportions correct over all questions.

|  | Humans | GPT-4 .5 hedge | GPT-4 0 hedge |
| --- | --- | --- | --- |
| Sample size | 1600 | 88 | 88 |
| Successes | 1489 | 85 | 82 |
| Proportion | .9305 | .9659 | .9318 |
| $Z$ |  | 1.2841 | .0429 |
| $p$ |  | .1991 | .9658 |

Note: hedged responses are scored as .5 or 0 in GPT-4.

Figure 1 shows the proportions correct in each of the four cells of the experiment (2 directness levels x 2 salience levels) for humans on the left and GPT-4 on the right. The overall pattern of proportions correct on the Discourse Comprehension Test (10) for GPT-4 closely resembles that for humans. Average neurotypical humans do very well on this test (12) while GPT-4 slightly exceeds human performance overall and in three of the four experimental cells portrayed in Figure 1. The pattern of proportions correct are roughly similar for humans and GPT-4 across the four experimental cells. Notably, the worst-performing cell for both humans and GPT-4 is the implied details cell.

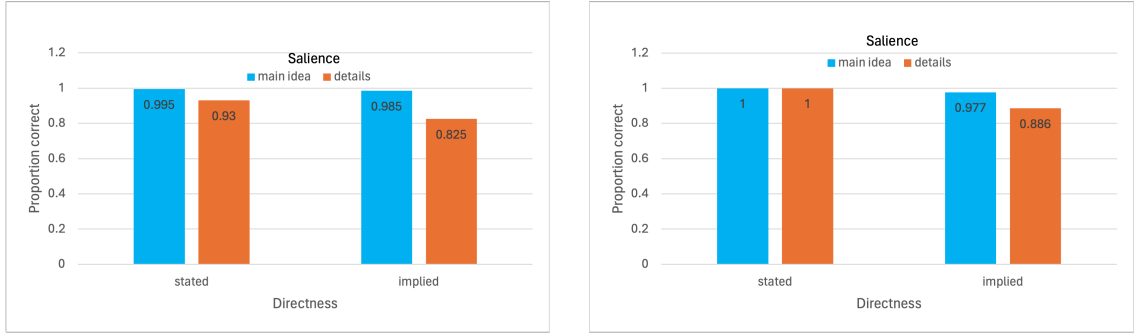

Figure 1. Proportions correct on the Discourse Comprehension Test for humans on the left and GPT-4 on the right, as a function of directness and salience of information.

For completeness, we assess whether humans and GPT-4 are performing better than chance, again using the Two Sample Independent Proportions Test. Here, chance performance is defined by .5 of sample sizes. The $Z$ and $p$ values in Table 2 certify that both neurotypical humans and GPT-4 models indeed perform well above chance.

Because of the theoretical interest in understanding of discourse via implication that goes beyond stated information, we compare GPT-4 to humans on stated-information questions (Table 3) and implied-information questions (Table 4). These comparisons use the slightly preferred scoring scheme that rates hedged responses as worth .5, as in Figure 1. Again, although GPT-4 does slightly better than humans on both stated and implied question information, the differences in each case are far from reaching statistical significance.

Table 2: Comparison of human and GPT-4 performance to chance, defined as .5 success.

|  | Humans | GPT-4 .5 hedge | GPT-4 0 hedge |
| --- | --- | --- | --- |
| Sample size | 1600 | 88 | 88 |
| Successes | 800 | 44 | 44 |
| Proportion | .9305 | .9659 | .9318 |
| $Z$ | 26.99 | 6.985 | 6.351 |
| $p$ | 0.0000 | 0.0000 | 0.0000 |

Table 3: Comparison of proportions correct on stated-information questions.

|  | Humans | GPT-4 |
| --- | --- | --- |
| Sample size | 800 | 44 |
| Successes | 770 | 44 |
| Proportion | .9625 | 1 |
| $Z$ | 1.3080 | |
| *p* | | .1909 |

Table 4: Comparison of proportions correct on implied-information questions.

| | Humans | GPT-4 |
|---|---|---|
| Sample size | 800 | 44 |
| Successes | 724 | 41 |
| Proportion | .9050 | .9315 |
| *Z* | | .5946 |
| *p* | | .5521 |

It is telling that GPT-4 never makes a wrong response in this experiment. As noted, it fails to give a *yes* or *no* response only 6 times out of 88, once on an implied main idea and five times on implied details. It hedges on each of these six cases, instead giving neutral uncertain responses and appropriate comments that justify their uncertainty.

We also experiment with GPT-4's ability to summarize these stories, finding that they produce a concise and accurate paragraph without much in the way of inferences. However, if we ask for a summary that mentions inferences, this opens the inferential floodgates. With that prompt, GPT-4 produces a total of 54 new inferences that go well beyond those used in the yes/no questions. The mean number of such inferences per story is 4.91, with a standard deviation of 2.02, and a range of 2 to 8. An example is provided in Appendix C, using the story presented in Appendix A.

2.3 Discussion

Our results show that GPT-4 matches and even slightly exceeds the high level of human performance on the Discourse Comprehension Test (10). Due to excellent human performance, there is very little room for GPT-4 to exhibit superiority over humans.

It makes sense that the worst performance in both humans and GPT-4 is in the experiment cell for details and implied knowledge. With memory constraints, details may be ignored or forgotten in favor of main points. And producing implications requires additional cognitive effort.

We encourage readers to carefully consider the example story presented throughout Appendices A, B, and C. The combination of never giving a wrong answer while spontaneously providing explanatory justifications makes it hard to believe that a story is not well understood by GPT-4. The same impression is given by GPT-4's spontaneous comments about questions in each of the other ten stories.

We are unable to suppress hedging and comments from GPT-4. Its comments on this task are both appropriate and interesting, often justifying a *yes*-or-*no* answer and sometimes referring precisely to the process of implication. Number of comments across the eleven stories range from 0-8, with a mean of 3.27. Only one story generated no comments. Human comments were not recorded beyond their yes/no responses (12).

GPT-4's strong overall performance on these novel stories suggests that it indeed understands what it has just learned in a single shot, even when that requires inferencing beyond what is directly stated in the story.

Because inferences are required to comprehend most if not all discourses (13), it is very likely that there is already considerable evidence in the LLM literature that GPT-4 uses inference in understanding what it reads (3,14). What is unique about our study is the deliberate experimental separation of salience and directness. This enables focusing more precisely on how these two important variables operate and interact. Fortuitously, the existence and use of the Discourse Comprehension Test provides data allowing a close human comparison while maintaining this clear separation between important variables on the same content.

Classical psychology experiments on discourse comprehension typically gave participants a paragraph to read and then asked them to write down what they remembered of the paragraph (15,16). The experimenter would then count the number of correctly recalled propositions as a measure of understanding. For several reasons, this methodology did not provide many interesting insights into discourse comprehension. It confounded understanding with memory, made no distinction between stated and implied information, and generally ignored the important role of inference based on general knowledge. In contrast, the Discourse Comprehension Test (10) separates direct from implied information and GPT-4 supplies extensive general world knowledge that promotes interesting and useful inferences.

A close analog to asking a human participant to write out a remembered paragraph is to ask GPT-4 to summarize what it has just read. This results in a very concise summary with little or no hint of inferencing. However, as noted in the 2.2 Results section, when we request GPT-4 to mention inferences to accompany their concise story summary, we discover that it provides many inferences that go well beyond the modest inferencing apparent in our first experiment with yes/no questions. It might be interesting to see whether human participants would likewise provide additional inferences if similarly prompted in this task.

3. Understanding More Difficult Passages

The lack of statistically significant differences between humans and GPT-4 in section 2 could be due to the relative simplicity of the stories used in the Discourse Comprehension Test (10). Both classes of participants performed at a sufficiently high level that there was very little room for one type to statistically exceed the performance of the other type. Our preliminary conclusion is that GPT-4 at least matched human performance on discourse comprehension. Here in section 3, we use considerably more difficult reading material, to allow greater possible distance between humans and GPT-4 in one direction or the other.

3.1 Overall Test Results

Large teams of OpenAI researchers recently published an extensive and detailed Technical Report on the capabilities, limitations, and safety characteristics of GPT-4 (17). Among the capabilities that they addressed were performances on 34 academic tests covering a wide range of fields. Three of these academic tests had sections that addressed reading comprehension at higher levels of difficulty than the Discourse Comprehension Test used in our section 2: SAT, GRE, and LSAT.

Our section 3.1 is a review of GPT-4 performance on these three widely used and highly standardized academic tests (17). They each have a large component devoted to reading comprehension. OpenAI researchers verified that there was no special GPT-4 training for these three tests, and they also ran contamination checks for test data appearing in the training set (17). If matches to the test set were found in the training set, they were removed from the test set to create an uncontaminated test set.

Table 5 shows the percentile achieved by GPT-4 in each test after eliminating any contamination from the training set. The mean uncontaminated percentile across the three tests is 96.3. By statistical definition, the average percentile achieved by thousands of student test-takers is the 50$^{th}$ percentile, thus revealing a substantial superiority for GPT-4 with reading comprehension of difficult passages. The prompts given to GPT-4 reflected the test requirements (17).

Table 5. GPT-4 Uncontaminated Percentile Scores on 3 Academic Tests that include Reading Comprehension.

| Test | Percentile |
|---|---|
| Scholastic Aptitude Test (SAT) Reading & Writing | 93$^{rd}$ |
| Graduate Record Examination (GRE) Verbal | 99$^{th}$ |
| Law School Admission Test (LSAT) | 97$^{th}$ |

Adapted from OpenAI 2024, their Table 9 in their Appendix G. (17)

For the SAT and GRE Verbal exams, scores were identical with and without contamination, while for the LSAT, GPT-4 performed slightly better on uncontaminated questions. This finding supports OpenAI's conclusion that contamination had little to no effect on GPT-4's scores and suggests that GPT-4's high scores reflect its reading comprehension abilities rather than specific memorized content from training data (17).

The SAT is widely used for college admissions in North America. The Reading section has brief passages (or a passage pair) followed by a multiple-choice question. Passages range from 25 to 150 words. The subject areas for Reading and Writing cover literature, history, social studies, humanities, and science. Students have 64 minutes to complete the Reading and Writing section.

Reading Comprehension questions on the GRE are designed to test for the ability to understand the kinds of prose commonly encountered in graduate and professional schools, including drawing conclusions from information, reasoning from incomplete data to infer missing information, understanding how the parts of a passage relate to each other, analyzing a text and reaching its conclusions, considering alternative explanations, and formulating and testing hypotheses. Test passages are borrowed from academic and non-academic books and articles covering science, arts, humanities, business, and everyday topics.

Reading comprehension passages and questions on the LSAT seem particularly well suited to discovering indications of true understanding as they often require the reader to reason beyond the literal text. Their multiple-choice questions probe for main ideas, explicitly stated information, inferable information, generalization to different contexts, and analogizing.

3.2 Other Signatures of Understanding

Although there are no precise experimental distinctions in these academic tests between stated and inferred information and between main points and details, as in The Discourse Comprehension Test (10), it is still possible to identify important signatures of text understanding such as generalization and inference. Our next step was to probe this more deeply by running a GPT-4 experiment with online available LSAT passages which were accompanied by explanations for the correctness and incorrectness of multiple-choice responses. Human students could read the correctness information for each multiple-choice answer as a useful pre-test study guide, while we instead prompt GPT-4 to provide justification for each of its answers in the test. Providing justifications is quite different and more challenging than the OpenAI testing which more closely matched the testing conditions for LSAT student test takers where justifications were not requested (17).

3.3 Method for probing LSAT performance more deeply

Our simulation experiment presents GPT-4 with three single passages and a pair of two related passages. For each of these four events, the generic prompt to GPT-4 is "Read this passage and then answer the two multiple-choice questions that follow. Also justify your answer to each question." The number of questions mentioned in the prompt varies from two to eight. The four test passages concern Roy Lichtenstein's pop art (3 questions), indigenous rights in Canada (2 questions), an alleged speculative bubble in tulip prices in the seventeenth-century Dutch tulip market (2 questions), and the extent of human involvement in climate change (pair of two passages, 7 questions). This LSAT prep test was arbitrarily chosen from several such LSAT prep tests available online. The simulation was performed with Copilot GPT-4 on 31 May 2024.

3.4 Results

The result is that GPT-4 gets all fourteen questions correct, approximately consistent with OpenAI's 97th percentile GPT-4 performance on entire LSATs (17). To examine GPT-4's cognitive performance in more detail, we display here the speculative-bubble passage as a single, but representative, example. In blue font are the prompt, passage, and questions. We encourage our readers to read this passage and then quickly answer multiple-choice questions 6 and 7 before reading the answers and explanations supplied by GPT-4 (below) or the test-maker (in Appendix D). This would provide a good idea of what the students and GPT-4 were up against in the LSAT.

Read this passage and then answer the two multiple-choice questions that follow. Also justify your answer to each question.

In economics, the term "speculative bubble" refers to a large upward move in an asset's price driven not by the asset's fundamentals—that is, by the earnings derivable from the asset—but rather by mere speculation that someone else will be willing to pay a higher price for it. The price increase is then followed by a dramatic decline in price, due to a loss in confidence that the price will continue to rise, and the "bubble" is said to have burst. According to Charles Mackay's classic nineteenth-century account, the seventeenth-century Dutch tulip market provides an example of a speculative bubble. But the economist Peter Garber challenges Mackay's view, arguing that there is no evidence that the Dutch tulip market really involved a speculative bubble.

By the seventeenth century, the Netherlands had become a center of cultivation and development of new tulip varieties, and a market had developed in which rare varieties of bulbs sold at high prices. For example, a Semper Augustus bulb sold in 1625 for an amount of gold worth about U.S. $11,000 in 1999. Common bulb varieties, on the other hand, sold for very low prices. According to Mackay, by 1636 rapid price rises attracted speculators, and prices of many varieties surged upward from November 1636 through January 1637. Mackay further states that in February 1637 prices suddenly collapsed; bulbs could not be sold at 10 percent of their peak values. By 1739, the prices of all the most prized kinds of bulbs had fallen to no more than one two-hundredth of 1 percent of Semper Augustus's peak price.

Garber acknowledges that bulb prices increased dramatically from 1636 to 1637 and eventually reached very low levels. But he argues that this episode should not be described as a speculative bubble, for the increase and eventual decline in bulb prices can be explained in terms of the fundamentals. Garber argues that a standard pricing pattern occurs for new varieties of flowers. When a particularly prized variety is developed, its original bulb sells for a high price. Thus, the dramatic rise in the price of some original tulip bulbs could have resulted as tulips in general, and certain varieties in particular, became fashionable. However, as the prized bulbs become more readily available through reproduction from the original bulb, their price falls rapidly; after less than 30 years, bulbs sell at reproduction cost. But this does not mean that the high prices of original bulbs are irrational, for earnings derivable from the millions of bulbs descendent from the original bulbs can be very high, even if each individual descendent bulb commands a very low price. Given that an original bulb can generate a reasonable return on investment even if the price of descendent bulbs decreases dramatically, a rapid rise and eventual fall of tulip bulb prices need not indicate a speculative bubble.

Question 6

The phrase "standard pricing pattern" as used in the middle of the last paragraph most nearly means a pricing pattern

    A. against which other pricing patterns are to be measured

    B. that conforms to a commonly agreed-upon criterion

    C. that is merely acceptable

    D. that regularly recurs in certain types of cases

    E. that serves as an exemplar

Question 7

Given Garber's account of the seventeenth-century Dutch tulip market, which one of the following is most analogous to someone who bought a tulip bulb of a certain variety in that market at a very high price, only to sell a bulb of that variety at a much lower price?

A. someone who, after learning that many others had withdrawn their applications for a particular job, applied for the job in the belief that there would be less competition for it

B. an art dealer who, after paying a very high price for a new painting, sells it at a very low price because it is now considered to be an inferior work

C. someone who, after buying a box of rare motorcycle parts at a very high price, is forced to sell them at a much lower price because of the sudden availability of cheap substitute parts

D. a publisher who pays an extremely high price for a new novel only to sell copies at a price affordable to nearly everyone

E. an airline that, after selling most of the tickets for seats on a plane at a very high price, must sell the remaining tickets at a very low price

Here are GPT-4's responses, also in blue font:

**Question 6** The correct answer is **D. that regularly recurs in certain types of cases**. The passage explains that Garber argues a "standard pricing pattern" occurs for new varieties of flowers, where the price of a particularly prized variety is initially high but falls rapidly as the bulbs become more readily available through reproduction. This pattern is not described as an exemplar (E), a commonly agreed-upon criterion (B), or a measure against which other patterns are compared (A). It's also not described as merely acceptable (C). Instead, it's a pattern that regularly recurs in the specific case of new flower varieties.

**Question 7** The correct answer is **D. a publisher who pays an extremely high price for a new novel only to sell copies at a price affordable to nearly everyone**. This is analogous to buying a tulip bulb of a certain variety at a very high price and then selling bulbs of that variety at a much lower price, as the original bulb can generate many descendent bulbs. Just like a publisher who pays a high price for a new novel (the original "bulb") and then sells many copies (the "descendent bulbs") at a lower price. The other options do not capture this dynamic of an initial high-cost investment that can generate many lower-cost items.

These justifications from GPT-4 suggest that it is, among other things, reasoning by generalization to other examples of price changes that conform to the standard pricing pattern of investing a lot in what people want and eventual selling it in high volume at a low price. For example, a publishing company pays a lot for a book which can then be sold to many readers at a much lower price. Such generalization strengthens explanations that apply more widely and are more correct and more useful. Generalization is a standard inferential signature of understanding textual discourse.

Generalization was also evident in two other of our four passages: pop art and climate change. There is a question about Lichtenstein's motivation for doing pop art in the way he did. Because motivation is not mentioned in the passage, the reader must generalize across the author's comments about Lichtenstein's approach to art, his rebellion against abstract expressionism, his incorporation of realism and naiveté, and his depiction of contemporary life.

In the two climate-change passages, global warming is attributed to human activities in passage A and to natural cycles in Passage B. In each of the two passages, there is a generalized explanation of several contemporary extreme weather phenomena, pitting one generalized explanation against another and enabling correct answers to several questions.

In the passage on indigenous rights in Canada, there is an important inference that indigenous oral tradition is necessary for establishing legal documentation of land claims. This is because the only cultural customs being considered occurred prior to the establishment of British sovereignty over the specific territories. Relevant written documentation would become available only after colonization.

We noticed in this experiment that GPT-4 explains its responses in far fewer words than the test makers used to explain the correctness of answers to students trying to ace their test. The test-maker explanations are available in Appendix D, where readers can judge for themselves which explanations they would prefer. We prefer the more concise explanations given by GPT-4 because they are just as informative and easier to understand compared to the wordier test-maker explanations. Using too many words in an explanation stresses memory and makes understanding more difficult. As we noted in the introduction, making good summaries of text is considered an indicator of understanding.

Many of the words used in test-maker explanations were used to explain why each of the four wrong answers were not correct. Even if we remove all the words addressing wrong answers from the statistical analysis, there is still a very strong tendency for GPT-4's explanations to be more concise than those of the test-makers, $t(13) = 7.48$, $p < .0001$, as shown in Figure 2. The 95% confidence interval of the mean difference is from 102.81 to 186.34.

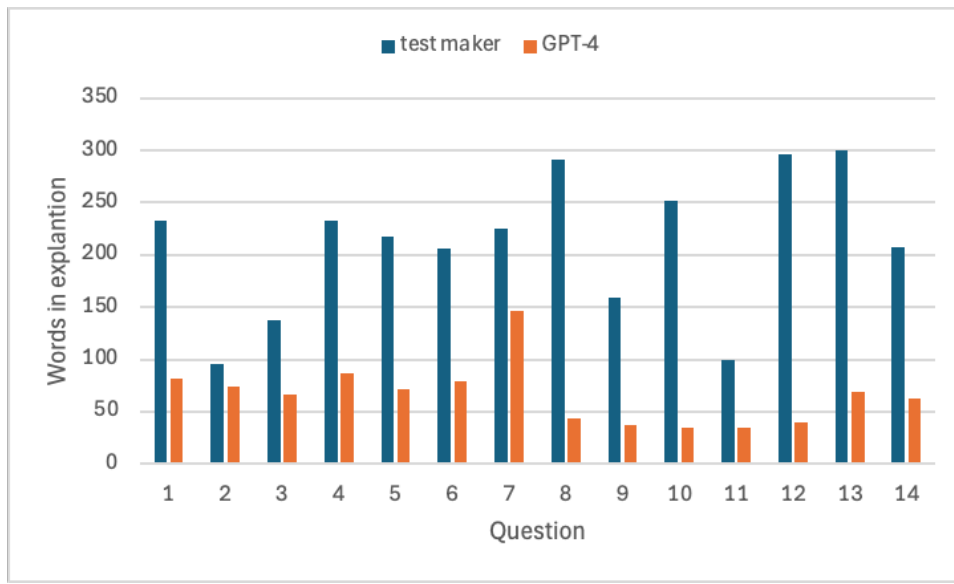

Figure 2: Number of words used to explain answer correctness.

It is likely that some of the extra words used by the test makers are useful in some way. Nonetheless, it is clear from these data that the more concise justifications provided by GPT-4 are sufficient to support perfect performance on this test by GPT-4, and that the more verbose explanations supplied by the test makers are not sufficient to produce perfect performance in human test takers, as the average human performance on LSATs falls far below the 100[th] percentile achieved here by GPT-4.

4. General Discussion

We report in section 2 that GPT-4 matches the performance of average adult humans on the Discourse Comprehension Test (10). This is notable on its own, but there is more to say about this phenomenon. Because the stories in that test are rather simple (5$^{th}$ and 6$^{th}$ grade reading levels), both humans and GPT-4 perform at a very high level. This raises the possibility that there is not sufficient room for one type of participant to perform at a higher level than the other type. We find in section 3.1 that increasing the difficulty of the text enables greater separation of the two subject types. GPT-4 here performs at a much higher level than do the humans for whom the more difficult tests were designed, i.e., highly motivated students striving to extend their education by doing well on admission tests. Performance differences on these more difficult passages and test questions are in the general neighborhood of 2:1 in favour of GPT-4 on the percentile scale. This provides substantial evidence that increasing the difficulty of text passages creates a strong interaction with participant type. Average humans do about as well with understanding simple text as does GPT-4, but GPT-4 can greatly exceed the performance of intelligent humans on more difficult passages.

Our converging evidence for genuine understanding of text in GPT-4 is strong due to a high level of correct responding to questions about the text, correct and concise justifications for those answers, and subtle signatures of understanding that go beyond the literal text such as generalization inferences and other inferences. It suggests that GPT-4 can exceed human levels of text understanding, provided that the text is sufficiently difficult.

Does this mean that GPT-4 has achieved Artificial General Intelligence (AGI)? It does not, because reading comprehension is only one skill in the AGI sphere, albeit a critically important skill because reading is the principal way that an LLM currently receives its knowledge. Nonetheless, reading comprehension can now be added to the long and growing list of cognitive skills on which GPT-4 has matched or exceeded average human performance: decision making (6), metaphors, idioms, estimation, anaphoric reference, creativity, problem solving, and answering common sense questions (3).

However, as we noted in section 1, LLMs still have difficulty with causal reasoning (6), planning (7), and graduate-level mathematics (8). Causal reasoning might well require models of agentive interaction with the physical world, involving the control of purposeful sensorimotor interactions (18). As well, a case could be made that such factors should be employed in AI construction well before the agents are capable of reading. Even very young infants act as though they are aware that their actions cause certain effects (19,20). Although infants are far from being able to read, their causal knowledge gets firmly grounded in purposely acting on the world to produce desired effects.

Another promising approach to causal reasoning involves using LLMs to provide prior probabilities to construct causal graphs (21). Such information from LLMs was found to improve performance on commonsense causal benchmark phenomena, especially when deciding what is the cause and what is the effect.

One thing to keep in mind when considering the high percentiles achieved by GPT-4 on difficult academic tests is that there are a small number of human test-takers who achieved even higher percentiles than achieved by GTP-4 on those tests. It could be interesting to study such people while monitoring the ongoing quest of LLMs striving for perfect success on important tasks. We happened to encounter some aspects of this issue in our final experiment using a small 14-question snippet of an LSAT test on which GPT-4 got every question correct. Some of those questions were flagged as very difficult by test-makers based on their difficulty for humans.

Although our article is presented in a kind of competition between humans and AI, we believe that LLMs could rapidly become of great service to interested humans. For example, humans could come to rely on yet to be fully developed AI systems that can read widely and reliably explain to us what they read and understand.

Acknowledgments

This research is supported by a grant from the Natural Sciences and Engineering Research Council of Canada to TRS: NSERC 2021-03434.

Appendices

Appendix A: Example story and its questions, with their correct answers and values of the salience and directness variables.

Appendix B: Responses of GPT-4 to the story in Appendix A.

Appendix C: Responses of GPT-4 to the story in Appendix A when prompted for inferences, which are highlighted by us in square brackets.

Appendix D: Test-maker explanations of answer correctness in Section 3.

Appendix A: Example story and its questions, with their correct answers and values of the salience and directness variables in Section 2.

Read this story in preparation for answering eight yes/no questions about the story.

One day last Fall, several women on Willow Street decided to have a garage sale. They collected odds and ends from all over the neighborhood. Then they spent an entire day putting prices on the things that they had collected. On the first day of the sale, they put up signs at both ends of the block and another one at a nearby shopping center. Next, they made a pitcher of iced tea and sat down in a shady spot beside the Anderson's garage to wait for their first customer. Soon a man drove up in an old truck. He looked around and finally stopped by a lumpy old mattress that was leaning against the wall. He gestured to it and asked how much they wanted for it. Mrs. Anderson told him that it wasn't for sale. Then she added that they were going to put it out for the trash collectors the next day. The man asked if he could have it. Mrs. Anderson said that he could. Then she asked, "Why do you want such a terrible mattress?" "Well," he said, "My no-good father-in-law is coming to visit next week, and I don't want him to get too comfortable."

1. Did several women have a party? No, stated main idea.

2. Were there a large number of things at the garage sale? Yes, implied main idea.

3. Did the women put up a sign at a shopping center? Yes, stated detail.

4. Was it cold the day of the garage sale? No, implied detail.

5. Was the man driving a car? No, stated detail.

6. Was the mattress in terrible condition? Yes, stated main idea.

7. Was the man married? Yes, implied detail.

8. Was the man fond of his father-in-law? No, implied main idea.

Appendix B: Responses of GPT-4 to the story in Appendix 1.

Note that we add parentheses around the spontaneous elaborations given by GPT-4 beyond its yes/no answers, which are printed in bold font by GPT-4.

Answer each of these yes/no questions about the story.

1. Did several women have a party? **No**
2. Were there a large number of things at the garage sale? **Yes**
3. Did the women put up a sign at a shopping center? **Yes**
4. Was it cold the day of the garage sale? **No** (The story does not provide information about the temperature)
5. Was the man driving a car? **No** (He drove up in an old truck)
6. Was the mattress in terrible condition? **Yes**
7. Was the man married? **Yes** (He mentioned his father-in-law)
8. Was the man fond of his father-in-law? **No** (He referred to his father-in-law as "no-good" and didn't want him to get too comfortable)

Appendix C: Responses of GPT-4 to the story in Appendix 1 when prompted for inferences, which are highlighted by us in square brackets.

Summarize this story, mentioning main ideas that are not stated and must be inferred.

Certainly! Here's a concise summary of the paragraph:

Several women on Willow Street organized a garage sale during the fall. They gathered various items from the neighborhood, priced them, and set up signs for the sale. On the first day, they sat down near the Anderson's garage, waiting for customers. A man arrived in an old truck and showed interest in a lumpy old mattress leaning against the wall. Mrs. Anderson informed him that it wasn't for sale and would be discarded.

Certainly! Let's delve into the main ideas that can be inferred from this delightful story:

1. **Community Spirit and Resourcefulness**:

    1.1. The women on Willow Street come together to organize a garage sale, [demonstrating their community spirit].

    1.2. They collect various items from their neighborhood, [showcasing resourcefulness and a desire to repurpose things].

2. **Preparation and Effort**:

    2.1. The women spend an entire day pricing the collected items, [emphasizing their dedication and hard work].

    2.2. They strategically place signs at both ends of the block and near a shopping center to attract customers, [highlighting their thoughtful preparation].

3. **Unexpected Interaction**:

    3.1. A man arrives in an old truck, showing interest in a lumpy old mattress leaning against the wall.

    3.2. Mrs. Anderson informs him that the mattress isn't for sale and will be discarded.

    3.3. The man's [request to take the mattress leads to an unexpected interaction].

4. **Humorous Twist**:

    4.1. When Mrs. Anderson questions why he wants such a terrible mattress, the man reveals his motive: his "no-good" father-in-law is visiting next week, and he doesn't want him to get too comfortable.

    4.2. [The humorous twist lies in using the uncomfortable mattress as a subtle way to inconvenience the unwelcome guest].

5. Overall, the story highlights camaraderie, practicality, and a touch of wit in dealing with unwanted family visits.

Appendix D**:** Test-maker explanations of answer correctness in Section 3.

We believe that the test-maker explanations are less effective than GPT-4's justifications because GPT-4 is much more concise. See Figure 2 in the main text.

Question 6
The phrase "standard pricing pattern" as used in the middle of the last paragraph most nearly means a pricing pattern
- A. against which other pricing patterns are to be measured
- B. that conforms to a commonly agreed-upon criterion
- C. that is merely acceptable
- D. that regularly recurs in certain types of cases
- E. that serves as an exemplar

Explanation for Question 6

This question requires the test taker to understand from context the meaning of the phrase "standard pricing pattern," which is used by the author in a particular way.
The correct answer choice is (D). The phrase occurs in the last paragraph of the passage. The purpose of this paragraph is to detail Garber's reasons for thinking that, contrary to Mackay's view, the seventeenth-century Dutch tulip market did not involve a speculative bubble. It is in this context that the author uses the phrase in question. The complete sentence reads, "Garber argues that a standard pricing pattern occurs for new varieties of flowers." The author then explains this standard pricing pattern: original bulbs for prized new varieties initially command a high price, but descendants produced from the original bulbs cost dramatically less over time. It is clear that the author takes Garber to be describing a regularly recurring pattern about the pricing of new varieties of flowers, and then asserting that the particular details about the pricing of tulip bulbs in the seventeenth century fit this recurring pattern. Thus, answer choice (D) is correct, since it paraphrases the use of the term "standard pricing pattern" as a pricing pattern "that regularly recurs in certain types of cases."

Answer choice (A) is incorrect. Nowhere does the author suggest that pricing patterns can or should be "measured" against one another, much less against a pricing pattern that is for some reason taken to be the benchmark.

Answer choice (B) is incorrect. The passage as a whole does concern the interpretation of the pricing of tulip bulbs in the seventeenth century, and it might be said that the debate between Mackay and Garber concerns whether this case fits commonly agreed-upon criteria regarding speculative bubbles. However, in the middle of the last paragraph Garber's point is simply about prices fitting a pattern observed in a number of other cases. In this way, it is a point about conformance to a historical pattern, not to agreed-upon standards.

Answer choice (C) is incorrect. There is no reason to think that the author views pricing patterns as "acceptable" or unacceptable, or that the author believes there is a standard for acceptability.
Answer choice (E) is incorrect. An "exemplar" would be a particular case that serves as some kind of model or ideal. No particular case is being offered up as a model in the third paragraph. Instead the "standard pricing pattern" is only described generally, not by reference to some paradigm example of the pattern Garber has in mind.

Based on the number of test takers who answered this question correctly when it appeared on the LSAT, this was a difficult question.

Question 7

Given Garber's account of the seventeenth-century Dutch tulip market, which one of the following is most analogous to someone who bought a tulip bulb of a certain variety in that market at a very high price, only to sell a bulb of that variety at a much lower price?
   A. someone who, after learning that many others had withdrawn their applications for a particular job, applied for the job in the belief that there would be less competition for it
   B. an art dealer who, after paying a very high price for a new painting, sells it at a very low price because it is now considered to be an inferior work
   C. someone who, after buying a box of rare motorcycle parts at a very high price, is forced to sell them at a much lower price because of the sudden availability of cheap substitute parts
   D. a publisher who pays an extremely high price for a new novel only to sell copies at a price affordable to nearly everyone
   E. an airline that, after selling most of the tickets for seats on a plane at a very high price, must sell the remaining tickets at a very low price

Explanation for Question 7

This question requires the test taker to identify the scenario that is most analogous to the way in which Garber would view the purchase of a tulip bulb at a very high price, and the later sale of tulip bulbs of that same variety at a much lower price. Before looking at the answer choices, it is worth getting clear on the specifics of Garber's account. In Garber's view, the value of the original bulb reflects the earnings that can be made from the descendant bulbs. Since an original bulb will produce multiple descendants, the value of the original will be much greater than the value of any individual descendant. The value of the original reflects the cumulative value of the descendants. Thus, someone could buy an original bulb at a very high price and still turn a profit by selling descendant bulbs at a much lower price.

The correct answer choice is (D). The relation between the manuscript of a new novel and the copies that can be made of that novel is analogous to the relation between an original bulb and its descendants. From the original novel, the publisher can produce many copies. The copies can then be sold for a much lower price than the original. The value of the new novel reflects the cumulative value of the sales of the copies.

Answer choice (A) is incorrect. The scenario described does not include anything akin to the relationship between an original bulb and later descendants. Instead, it presents an example of someone who applies for a job based on a perception about the degree of competition for that job.

Answer choice (B) is incorrect. In this scenario, the value of the painting has dropped due to critical or public opinion. This represents a case in which the art dealer has taken a loss, not one where the art dealer recoups the original value of the painting through an accumulation of smaller sales.

Answer choice (C) is incorrect. On the surface, the drop in price of the motorcycle parts due to a flooded market of replacement parts seems similar to the drop in price of the bulbs of a variety of flowers. However, the situation is disanalogous in crucial respects. The cheap substitute parts cannot be described as anything like "descendants" of the original rare parts, and the owner of the box of rare parts does not get the value back through the cumulative sales of the cheap replacements. Indeed, the owner of the box of rare motorcycle parts was simply forced to sell the parts at a loss.

Answer choice (E) is incorrect. The airline had a certain number of seats for which they could sell tickets. The drop in price over time is not a product of increased availability, as in the case of the flower bulbs. In this case, the number of available seats has actually decreased. While it is surely rational for the airline to reduce the price of the seats, the situation is not analogous to the drop in price of descendant flower bulbs.

Based on the number of test takers who answered this question correctly when it appeared on the LSAT, this was a difficult question.